\documentclass{bmvc2k}
\usepackage{blindtext}
\usepackage{geometry}
\usepackage{booktabs}
\usepackage{graphicx}
\usepackage{multirow}
\usepackage{pifont}
\usepackage{arydshln}
\usepackage{placeins}
\usepackage{xcolor}

\title{Semantic Segmentation under Adverse Conditions: A Weather and Nighttime- aware Synthetic Data-based Approach}

\addauthor{Abdulrahman Kerim}{a.kerim@lancaster.ac.uk}{1}
\addauthor{Felipe Chamone}{cadar@dcc.ufmg.br}{2}
\addauthor{Washington Ramos}{washington.ramos@dcc.ufmg.br}{2}
\addauthor{Leandro Soriano Marcolino}{l.marcolino@lancaster.ac.uk}{1}
\addauthor{Erickson R. Nascimento}{erickson@dcc.ufmg.br}{2}
\addauthor{Richard Jiang}{r.jiang2@lancaster.ac.uk}{1}

\addinstitution{
 School of Computing and Communications\\
 Lancaster University\\
 Lancaster, UK
}
\addinstitution{
Computer Science Department\\
Universidade Federal de Minas Gerais,\\
Minas Gerais, Brazil
}

\runninghead{Kerim et al.}{Semantic Segmentation under Adverse Conditions}

\def\etal{\emph{et al}\bmvaOneDot}

\begin{document}

\maketitle

\begin{abstract}

Recent semantic segmentation models perform well under standard weather conditions and sufficient illumination but struggle with adverse weather conditions and nighttime. Collecting and annotating training data under these conditions is expensive, time-consuming, error-prone, and not always practical. Usually, synthetic data is used as a feasible data source to increase the amount of training data. However, just directly using synthetic data may actually harm the model's performance under normal weather conditions while getting only small gains in adverse situations. Therefore, we present a novel architecture specifically designed for using synthetic training data for domain adaptation. We propose a simple yet powerful addition to DeepLabV3+ 
% \en{add here a brief description/intuition about this simple and powerful addition}
by using weather and time-of-the-day supervisors trained with multi-task learning, making it both weather and nighttime aware, which improves its mIoU accuracy by $14$ percentage points on the ACDC dataset while maintaining a score of $75\%$ mIoU on the Cityscapes dataset. %\fc{[Cadar] I did some minor changes here}
Our code is available at \href{https://github.com/lsmcolab/Semantic-Segmentation-under-Adverse-Conditions}{https://github.com/lsmcolab/Semantic-Segmentation-under-Adverse-Conditions}.
% DeepLabV3+, HRNet, DANet, and PSPNet on our novel synthetic data
% \kr{mention that we propose new synthetic dataset AWSS and AWSS+}  
\end{abstract}

% Recent semantic segmentation models perform well under standard weather conditions and sufficient illumination but struggle with adverse weather conditions and nighttime. Collecting and annotating training data under these conditions is expensive, time-consuming, error-prone, and not always practical. Usually, synthetic data is used as a feasible data source to increase the amount of training data. However, just directly using synthetic data may actually harm the model's performance under normal weather conditions while getting only small gains in adverse situations. Therefore, we present a novel architecture specifically designed for using synthetic training data. We propose a simple yet powerful addition to DeepLabV3+ by using weather and time-of-the-day supervisors trained with multi-task learning, making it both weather and nighttime aware, which improves its mIoU accuracy by $14$ percentage points on the ACDC dataset while maintaining a score of $75\%$ mIoU on the Cityscapes dataset. 

% Kerim You should add these:
% 1. Qualitative Results.
% 2. Per-class results.
% 3. Failure cases for our method

%-------------------------------------------------------------------------
\section{Introduction}
\begin{figure*}[t]
    \centering
    \includegraphics[width=0.7\textwidth]{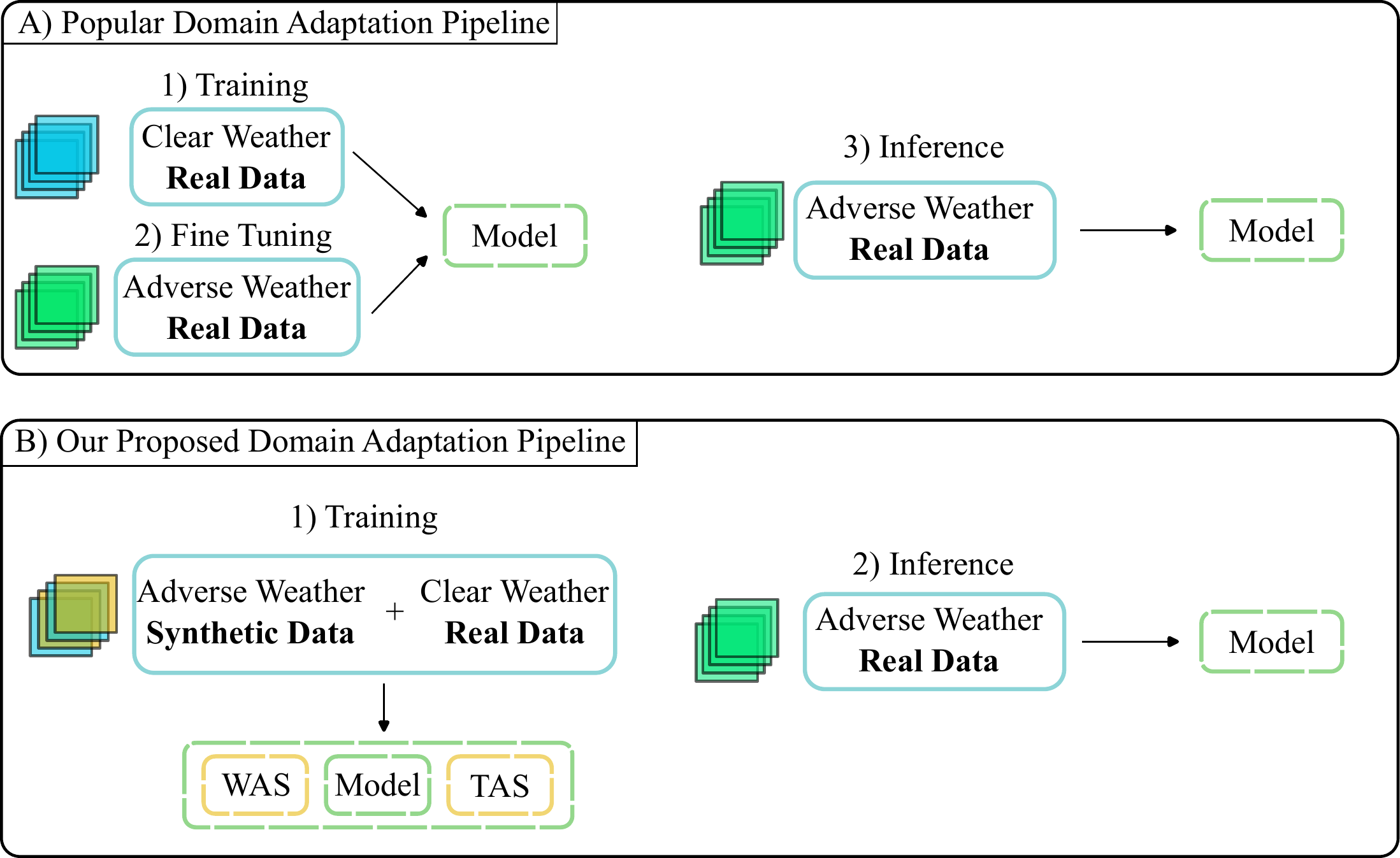}
    % \vspace{-18px}
    \caption{\textbf{Existing domain adaptation vs. our proposed pipeline.} Unlike other approaches, our pipeline utilizes synthetic data, Weather-Aware-Supervisor (WAS), and Time-Aware-Supervisor (TAS) to handle standard-to-adverse domain adaptation. Leveraging our synthetic-aware training procedure, we train our weather and daytime-nighttime aware architecture, simultaneously, on synthetic adverse weather and real normal weather data.}
    \label{fig:teaser}
    \vspace{-10px}
\end{figure*}

% === Establish the territory: semantic segmentation under adverse weather conditions ===
Understanding the environment using visual data has been an active research problem since the early beginning of computer vision. It started to attract even more researchers with the great advancement in autonomous cars~\cite{teichmann2018multinet,kumar2021syndistnet,wiseman2022autonomous}, human-computer-interaction~\cite{ren2020review,nazar2021systematic,liu2022human}, and augmented reality~\cite{baroroh2021systematic,costa2022augmented,chiang2022augmented}. Semantic segmentation is at the core of these applications, with the data-driven supervised learning methods dominating this field, achieving state-of-the-art results~\cite{ronneberger2015u,chen2018encoder,yuan2019segmentation,fu2019dual,zhao2017pyramid}. Training these models on real data requires large-scale human annotated images, which is expensive and time-consuming, especially for images taken under challenging weather and illumination conditions such as fog and nighttime. For instance, a person takes about $90$ minutes to annotate an image from the Cityscapes dataset~\cite{cordts2016cityscapes}, which contains only daylight and clear weather conditions, while it exceeds three hours for the Adverse Conditions Dataset with Correspondences (ACDC)~\cite{sakaridis2021acdc} dataset. %Due to this issue in the annotation efforts, the performance of most data-driven semantic segmentation models degrades under challenging weather and illumination conditions.

% === Establish the niche: semantic segmentation methods perform well under standard conditions but struggle under adverse weather conditions  ===
% what are the gaps in research that your study addresses?

Despite the success of recent semantic segmentation models in clear weather and standard illumination conditions, these methods struggle with adverse conditions (\textit{e.g.}, rainy, foggy, snowy, and nighttime), which degrade the feature extraction process. Falling rain and snow particles change the visual appearance of objects, partially occlude them, and cause distortion on the camera sensor, while fog works as a low-pass filter, removing high-frequency components. Nighttime is even more problematic because of the dramatic change in the light distribution and other severe artifacts, such as lens flare, bright spots, and chromatic aberration. Yet, few works have tried to investigate the effect of weather conditions and nighttime in semantic segmentation~\cite{guo2020high,lei2020semantic,alshammari2020competitive,xu2021cdada,ma2022both}. Although they achieve remarkable results, they are limited to one weather condition only and are too narrow in their scope.

% === Occupy the niche: your wonderful approach :-) ===
In this paper, we propose a novel training procedure to address the issues in the semantic segmentation under adverse conditions and in the annotation efforts, simultaneously. We leverage synthetic data to produce ground-truth images at no human annotation effort and create a new dataset, the AWSS, which is composed of images specially generated by a modified version of the \textit{Silver}~\cite{kerim2021silver} simulator. To reduce the gap between synthetic and real, our approach combines synthetic and real images by alternating their batches at training time as illustrated in Fig.~\ref{fig:teaser}. We also propose the Weather-Aware Supervisor (WAS) and the Time-Aware Supervisor (TAS), which are trained jointly with the main module to improve the feature extraction. Our main module derives from the DeepLabV3+ which contains the powerful atrous convolutions that increase the receptive field while not increasing the dimensions of feature maps and computation cost. Thus, better performance at low computation. Unlike the current methods that work only with a single weather condition, our approach can handle the three main ones, \textit{i.e.}, rainy, foggy, and snowy, as well as nighttime images. The results show that our novel model achieves state-of-the-art results under adverse weather conditions (0.49 mIoU on ACDC) while it maintains adequate performance under standard conditions (0.75 mIoU on Cityscapes).

%One primary objective is to show that evaluating segmentation methods on standard conditions, like the Cityscapes dataset~\cite{cordts2016cityscapes}, is not sufficient to assess robustness. At the same time, relying on real data is not viable as it is expensive, hard, and dangerous to capture under these challenging conditions. Labeling images acquired in such adverse conditions takes more time and effort and is more subjective to human error. It is hard for the human eye to segment objects when they are occluded, small-sized, and located in cluttered scenes. %These problems become even more acute under rainy, foggy, and snowy weather conditions. 
% Thus, real data, on its own, is not an optimal solution to train robust models.
%Therefore, we present a solution that can boost the robustness of recent models by utilizing specially generated synthetic data spanning these challenging conditions using a modified version of the \textit{Silver}~\cite{kerim2021silver} simulator. Additionally, we show that guiding the encoder of the segmentation model 
% \en{still not clear how is applied such guidance, i.e., what is the technical contribution?}
%\kr{by using weather and time-of-the-day supervisors and training them with multi-task learning using our synthetic-ware training procedure} help the model to learn weather and daytime-nighttime specific features, which significantly improves the performance under adverse conditions while still achieving adequate results under standard conditions. Our approach is illustrated in Figure~\ref{fig:teaser}.

In summary, our contributions are three-fold:
% \noindent \textit{i)} a new synthetic training data generator supporting semantic segmentation able to generate an unlimited number of training data under a wide set of challenging attributes such as adverse weather conditions and nighttime;
\noindent \textit{i)} a novel synthetic-aware training procedure %\en{you should include a brief description of this new training procedure in the previous paragraph --- it seems your technical contribution}
that can be used to train on both synthetic and real data simultaneously. In particular, we significantly improve DeepLabV3+~\cite{chen2018encoder} robustness on adverse conditions by making its encoder both weather and nighttime aware;\footnote{The synthetic data, code, and our modified version of the \textit{Silver}~\cite{kerim2021silver} simulator are all publicly available under the paper's GitHub repository.}
\noindent \textit{ii)} We extend the \textit{Silver}~\cite{kerim2021silver} simulator to generate more photo-realistic and diverse adverse weather conditions and increase the supported semantic segmentation classes;
\noindent \textit{iii)} leveraging our modified version of \textit{Silver}, we generate a new synthetic semantic segmentation dataset, the AWSS, composed of photo-realistic annotated images spanning foggy, rainy, and snowy weather conditions and nighttime attributes.

\begin{table}[t]
\centering
\caption{{\bf Comparison among synthetic semantic segmentation datasets.} Our dataset, named AWSS, is composed of photo-realistic pixel-wise annotated images under standard and adverse conditions.}
\label{table:synth_ds_comp}
\resizebox{\textwidth}{!}{%
\begin{tabular}{lcccccccc}
 &
  \multicolumn{4}{c}{Weather Conditions} &
  \multicolumn{2}{c}{Times-of-Day} &
  Photo-realism &
  \begin{tabular}[c]{@{}c@{}}Public\\ Availability\end{tabular} \\
\cmidrule(lr{1em}){2-5} \cmidrule(lr{1em}){6-7} \cmidrule(lr{1em}){8-8} \cmidrule(lr{1em}){9-9}
              & Normal     & Rain       & Fog        & Snow       & Daytime    & Nighttime  &      /      &     /        \\
\midrule
GTA-V~\cite{richter2016playing}          & \ding{52} & \ding{52} & -          & -          & \ding{52} & -          & \ding{52} & \ding{52} \\
Synscapes~\cite{tsirikoglou2017procedural, wrenninge2018synscapes}     & \ding{52} & -          & -          & -          & \ding{52} & -          & \ding{52}          & \ding{52} \\
Virtual KITTI~\cite{gaidon2016virtual} & \ding{52} & \ding{52} & \ding{52} & -          & \ding{52} & -          & -          & \ding{52} \\
Synthia~\cite{ros2016synthia}       & \ding{52} & \ding{52} & -          & \ding{52} & \ding{52} & \ding{52} & -          & -          \\
SHIFT~\cite{sun2022shift}       & \ding{52} & \ding{52} &  \ding{52}          & - & \ding{52} & \ding{52} &  \ding{52}          &  \ding{52}          \\
% \cdashline{1-2}
% \hdashline
AWSS (Ours) &
  \ding{52} &
  \ding{52} &
  \ding{52} &
  \ding{52} &
  \ding{52} &
  \ding{52} &
  \ding{52} &
  \ding{52}
\end{tabular}%
}
\end{table}

%------------------------------------------------------------------------
\section{Related Work}
%% Wash - I'm afraid we will not have room for this section introduction. I removed it for now.
%Our work utilizes synthetic data for domain adaptation  to improve performance under adverse conditions. We review the usage of synthetic data in the semantic segmentation field. At the same time, we highlight the recent progress in domain adaptation methods similar to our work. 

\paragraph*{Synthetic data for semantic segmentation.} 
The high performance of recent semantic segmentation models is associated with the ability to train deep models on large-scale training data. The early real semantic segmentation datasets like CamVid~\cite{brostow2009semantic}, Stanford Background~\cite{liu2010single, saxena2005learning, saxena2008make3d}, and KITTI-Layout~\cite{alvarez2012road} are limited in terms of the number of training samples, classes, resolution, and diversity. The problem is partially alleviated with the recent availability of datasets like Cityscapes~\cite{cordts2016cityscapes}, ACDC~\cite{sakaridis2021acdc}, ADE20K~\cite{zhou2017scene}, and Mapillary Vistas~\cite{neuhold2017mapillary}. Nevertheless, annotating large-scale datasets of high-resolution images is still the bottleneck. At the same time, ensuring diverse training data under challenging attributes like adverse weather conditions is not only dangerous, time-consuming, and hard to collect but also cumbersome and subjective to human errors in the annotation process. 

Synthetic data comes as a resort to handle all the above issues. Their success in computer vision is specifically seen in semantic segmentation. %Recent advancements in computer graphics motivated researchers to exploit synthetic data to overcome the effort of real data annotation and even improve semantic segmentation performance. 
%Using synthetic data complementary to the real data in the training phase improved the semantic segmentation performance. 
Goyal~\etal~\cite{goyal2017dataset} demonstrate that augmenting synthetic data with weakly annotated data can improve the performance on the PASCAL VOC dataset~\cite{everingham2015pascal}. Similarly, Richter~\etal~\cite{richter2016playing} generate synthetic training data by utilizing the Grand Theft Auto V game. They show that training semantic segmentation models on one third of the training split of CamVid~\cite{brostow2009semantic} dataset along with their generated synthetic data achieves superior results compared to training on the full CamVid~\cite{brostow2009semantic}. In parallel, Ivanovs~\etal~\cite{ivanovs2022improving} augment the Cityscapes~\cite{cordts2016cityscapes} dataset with synthetic images generated using the CARLA~\cite{dosovitskiy2017carla} simulator. They show that the performance improves when compared to training only on Cityscapes~\cite{cordts2016cityscapes}. Similar to these works, we use synthetic data to boost the performance of semantic segmentation models. However, we tackle the domain shift problem using synthetic data and a synthetic-aware training procedure. 

% \paragraph*{Using synthetic data for evaluation.}
% The usage of synthetic data was not limited to the training phase but also extended even for evaluation. While evaluating semantic segmentation models does not require a large-scale dataset, it does require a careful design. To accurately assess the continuous domain shift,    
% 1. Learning Semantic Segmentation from Synthetic Data:A Geometrically Guided Input-Output Adaptation Approach (check domain adaptation too)
% 2.X Improving Semantic Segmentation of Urban Scenes for Self-Driving Cars with Synthetic Images
% 3.X Playing for Data: ground-truth from Computer Games
% 4. Playing for Benchmarks
% 5. Play and Learn: Using Video Games to Train Computer Vision Models
% 6. The synthia dataset: A large collection of synthetic images for semantic segmentation of urban scenes
% 7. SHIFT: A Synthetic Driving Dataset for Continuous Multi-Task Domain Adaptation
% 8. Effective Use of Synthetic Data for Urban Scene Semantic Segmentation
% 9.X Dataset Augmentation with Synthetic Images Improves Semantic Segmentation
% \subsection{Semantic segmentation}

% Optimal foggy scene understanding – via adapting between two domains (normal and foggy) trained simultaneously with shared weights (each model is trained
% on one weather condition independently) and employing
% adversarial techniques on the output from each model.
% You can add snow and rain works?
\noindent\textbf{Domain adaptation in semantic segmentation.}
A major limitation of synthetic data is the domain shift: models trained on synthetic data do not perform well on real-world data~\cite{sankaranarayanan2018learning,xu2019self,dundar2020domain}. Sankaranarayanan~\etal~\cite{sankaranarayanan2018learning} propose a Generative Adversarial Network (GAN) based approach that minimizes the distance between the encodings of both domains. They show that their approach can boost the performance of synthetic-to-real domain adaptation tasks. Our work is similar to theirs as we use synthetic data for domain adaptation and propose a synthetic-aware training procedure. However, our work tackles this problem under harder set-ups utilizing synthetic data to mitigate standard-to-adverse domain shifts. In the same context, Alshammari~\etal~\cite{alshammari2020competitive} address standard to foggy weather domain shift by using an adversarial training strategy that guides the model to produce outputs close to the target domain. Similarly, Ma~\etal~\cite{ma2022both} tackle standard weather to foggy weather domain adaptation using both fog and style variations by adopting a Cumulative style-fog-dual disentanglement Domain Adaptation method (CuDA-Net). Alternatively, Xu~\etal~\cite{xu2021cdada} address the daytime to nighttime domain shift. They utilize a novel Curriculum Domain Adaptation method (CDAda) that uses labeled synthetic nighttime images. Our method is closely related to these works. However, we tackle domain adaptation from a standard domain (\textit{i.e.,} daytime and normal weather condition) to an adverse domain (\textit{i.e.}, nighttime and adverse weather conditions such as rain, fog, and snow).
%------------------------------------------------------------------------
\section{The AWSS Dataset}
There have been many synthetic datasets proposed for the semantic segmentation problem. However, they are usually non-photo-realistic such as Synthia~\cite{ros2016synthia} and Virtual KITTI~\cite{gaidon2016virtual}, limited in diversity such as GTA-V~\cite{richter2016playing} and Synscapes~\cite{tsirikoglou2017procedural, wrenninge2018synscapes} as clearly demonstrated in Table~\ref{table:synth_ds_comp}. Recently, SHIFT~\cite{sun2022shift} dataset was introduced, which is photo-realistic and diverse similar to our generated synthetic dataset but does not cover the snowy weather. 

\begin{figure*}[t]
    \centering
    \includegraphics[width=0.9\textwidth]{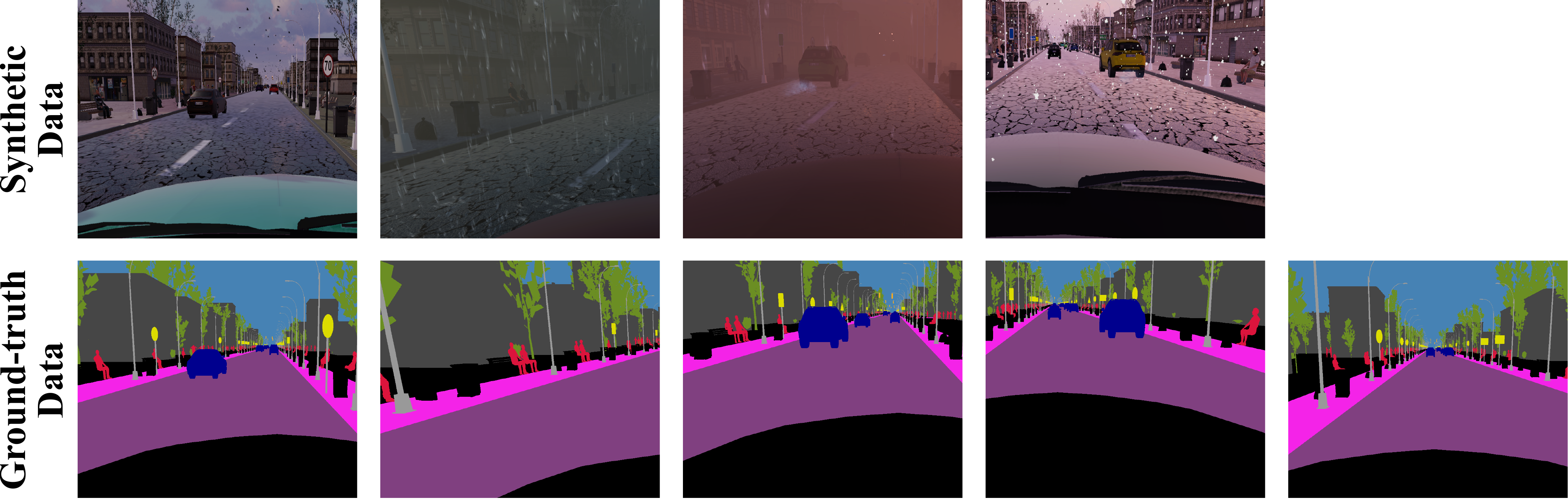}
    \caption{\textbf{Samples from AWSS dataset.} Our generated AWSS synthetic dataset spans normal, rainy, foggy, snowy, and nighttime attributes. }
    \label{fig:AWSS_diversity}
\end{figure*}

We extend \textit{Silver}, proposed by Kerim~\etal~\cite{kerim2021silver}, to generate adverse weather photo-realistic images along with their corresponding ground-truth for the semantic segmentation task. We generate the Adverse Weather Synthetic Segmentation (AWSS) dataset, which comprises $1{,}250$ images with a resolution of $1{,}200\times780$ pixels and spans normal, rainy, foggy, and snowy weather conditions at daytime and nighttime. It follows the same conventions, \textit{i.e.}, classes definitions and color encoding,
% \en{which ones?}
as Cityscapes~\cite{cordts2016cityscapes} and ACDC~\cite{sakaridis2021acdc} datasets. However, we limit the number of classes to $10$, namely \textit{Road}, \textit{Sidewalk}, \textit{Building},	\textit{Pole}, \textit{Traffic Light}, \textit{Traffic Sign}, \textit{Vegetation}, \textit{Sky}, \textit{Person}, and \textit{Car}. Figure~\ref{fig:AWSS_diversity} shows sample images from the AWSS dataset spanning various standard and challenging attributes.

% \subsection{Extensions to Silver framework}
\paragraph*{Extensions to the Silver framework.} \textit{Silver} is based on the Unity game engine~\cite{UnityGameEngine}. It allows users to create 3D virtual worlds by only specifying a set of scene descriptive parameters like the weather condition, time-of-the-day, number of cars and humans, camera type, and lens artifacts. The simulator achieves photo-realism by using the recent High Definition Rendering Pipeline (HDRP). In addition, the simulator applies a set of Procedural Content Generation (PCG) concepts to generate, populate, and control the scenes~\cite{kerim2021silver}. 

\noindent\textbf{i) Adverse conditions.}
The original simulator can simulate standard and adverse weather conditions
% : normal, rainy, foggy, and snowy weather conditions 
at daytime and nighttime but with a limited photo-realism and diversity. For each weather condition, we diversify weather severeness, time-of-the-day, and other scene elements if not specified. Based on the environment being simulated, scene elements materials, shaders and textures are selected from a predefined large set. We customize and integrate Procedural Terrain~\cite{ProceduralTerrain} with Adobe Substance materials~\cite{SubstanceMaterial} to simulate photo-realistic snow accumulation on ground, mud, mold, wet surfaces, and water puddles. Water drops splashes on the ground are simulated by customizing the Unity particle system. Rain splash intensity is controlled by the rain weather severeness which is sampled from a uniform distribution. Additionally, we simulate slightly foggy weather condition once heavy rain is simulated. For nighttime simulation, street lights are turned on and their intensity is randomized. Some of these lights are flickered or turned off to increase diversity.

\noindent\textbf{ii) Dash camera mode.}
Initially \textit{Silver} simulates Unmanned Aerial Vehicle (UAV) and first-person view cameras. However, most existing semantic segmentation datasets like Cityscapes~\cite{cordts2016cityscapes} and ACDC~\cite{sakaridis2021acdc} datasets are recorded using a dash camera mounted on a car. To generate our AWSS dataset, we develop the dash camera mode to facilitate this task. Furthermore, to increase view angle diversity, we simulate vertical and horizontal lens shifts.

\noindent\textbf{iii) Semantic segmentation automatic ground-truth.} 
The simulator supports semantic segmentation automatic ground-truth generation. However, the number of semantic classes was limited to 4: humans, ground, buildings, and trees. We extend the number of supported classes by adding new elements to the scene like traffic signs and modify the road mesh into road and sidewalk. At the same time, we customize the ground-truth generation pipeline to match Cityscapes~\cite{cordts2016cityscapes} color codes and conventions. With our extension, \textit{Silver} now can provide semantic segmentation ground-truth for 10 classes, as specified earlier in this section. %namely \textit{Road}, \textit{Sidewalk}, \textit{Building}, \textit{Pole}, \textit{Traffic Light}, \textit{Traffic Sign}, \textit{Vegetation}, \textit{Sky}, \textit{Person}, and \textit{Car}. 

\begin{figure*}[!t]
    \centering
    \includegraphics[width=\textwidth]{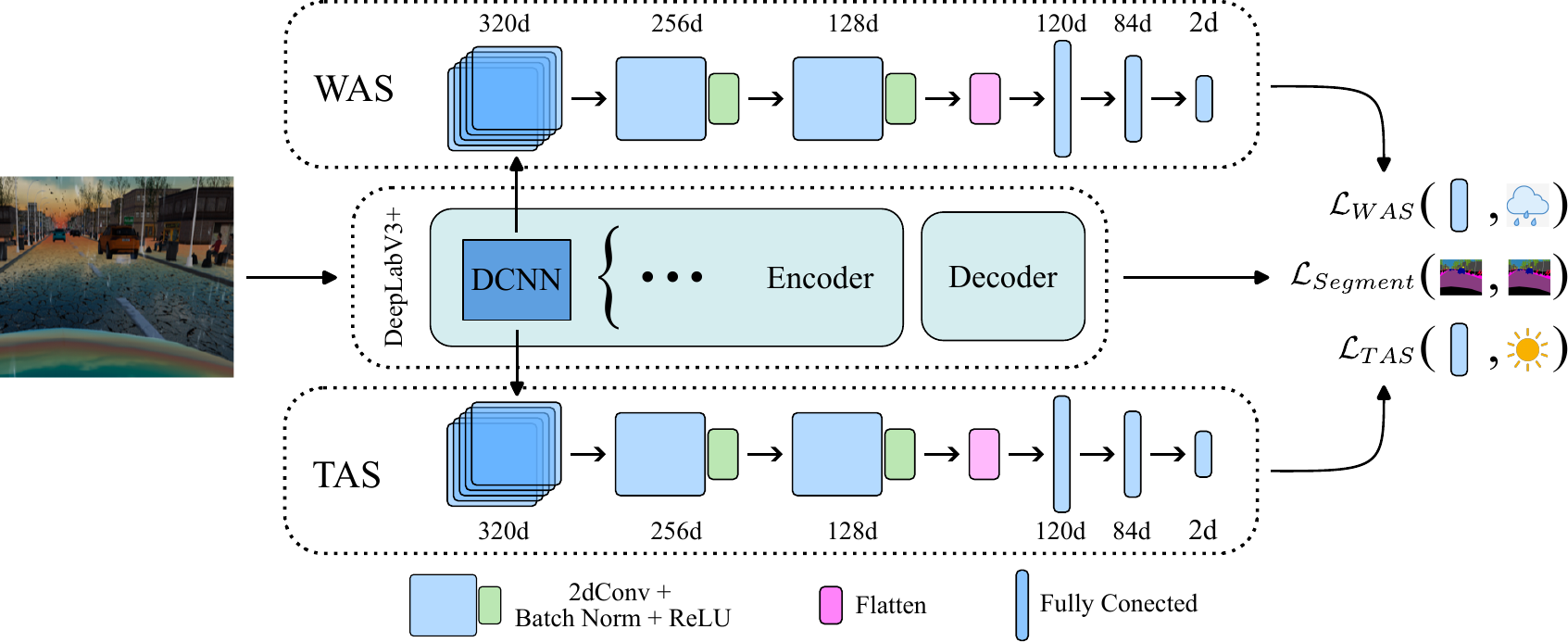}
    \caption{\textbf{An overview of our proposed architecture.} DCNN of DeepLabV3+~\cite{chen2018encoder} is forced to learn weather and daytime-nighttime specific  and roboust features by the means of multi-task learning. WAS and TAS branches learn to predict weather and daytime-nighttime, respectively. At the same time, they guide the encoder and specifically DCNN to learn extracting robust features under adverse and standard conditions. }
    \label{fig:methodology}
\end{figure*}

\section{Methodology}

% Our aim is to reduce the domain shift between normal and adverse weather domains while not acquiring any additional real data. Therefore, we propose a novel training approach that leverages synthetic data, and makes the architecture aware of the weather condition and time-of-the-day. % \en{Add here a brief description/introduction of your training approach that will be detailed in the following sections.} 
% \kr{Specifically, we propose training our aware architecture on data from both synthetic and real domains simultaneously and from scratch.  Our approach is illustrated in Figure~\ref{fig:methodology}.}

We aim to reduce the domain shift in adverse weather conditions while not acquiring additional real data. Hence, we propose a novel training approach that leverages synthetic data, while making the architecture aware of the weather condition and nighttime. Our architecture is trained on both synthetic and real data simultaneously (see Figure \ref{fig:methodology}). Our methodology is based on three components: i) adding two simple networks WAS and TAS that work as supervisors to teach the model to learn weather and nighttime specific features; ii) the full-model is trained using multi-task learning where the baseline learn semantic segmentation and WAS and TAS learns to predict weather condition and day-night, respectively; iii) the model is trained on images from synthetic domain~$\mathcal{D}_{adv-synth}$ and real domain~$\mathcal{D}_{stand-real}$ in alternating fashion to ensure that the model learn to extract adverse weather features only from synthetic data which presents a proxy of the adverse real domain~$\mathcal{D}_{adv-real}$. At the same time, it does not overfit to synthetic data and still ensure that the architecture other components leverage real data. Throughout the paper, $\mathcal{D}_{stand-real}$, $\mathcal{D}_{adv-real}$, and $\mathcal{D}_{adv-synth}$ are represented by Cityscapes~\cite{cordts2016cityscapes}, ACDC~\cite{sakaridis2021acdc}, and AWSS datasets, respectively. 
\noindent\textbf{Weather and nighttime aware encoder.}
% We utilize DeepLabV3+~\cite{chen2018encoder} architecture because of its powerful atrous convolution based encoder-decoder architecture making it a strong baseline. In the original architecture, it is assumed that the encoder will learn how to extract low-level and high-level features independent of the weather condition and time-of-the-day. This prevents the model from learning how to extract weather-specific features which result in low-quality (high-level and low-level) features being fed to decoder. Thus, clear degrading in performance under these conditions. The problem becomes even harder when there is no training samples under these conditions. 
We use the DeepLabV3+~\cite{chen2018encoder} architecture because of its powerful encoder-decoder architecture. Originally, it is assumed that the encoder will learn how to extract low-level and high-level features independent of weather and illumination conditions. This prevents the model from learning how to extract weather-specific features, resulting in low-quality features being fed to the decoder. The problem becomes even harder without training samples under these conditions.

% input is torch.Size([4, 320, 48, 48])
%  Tr image is 768x768x3
To alleviate this problem, we focus on the Deep Convolutional Neural Network (DCNN) which is a modified version of Xception~\cite{chollet2017xception}. We leverage multi-task learning to enforce the encoder to learn weather and time specific features. We add two simple identical models Weather-Aware-Supervisor (WAS) and Time-Aware-Supervisor (TAS). Each model is composed of two $3\times3$ atrous 2D convolutions with a rate of $2$ and padding of $6$. Each convolution is followed by a batch normalization and a rectified linear unit (ReLU). After this, the feature map is flattened and fed to 3 fully connected layers. The last layer predicts the weather for WAS and the daytime-nighttime for TAS. It is worth noting that WAS and TAS are only activated in the training process to guide the feature extraction learning process.   

\noindent\textbf{Multi-task learning to improve semantic segmentation.}
In the original implementation of DeepLabV3+~\cite{chen2018encoder}, the output of DCNN is passed to the remaining part of the encoder and to the decoder. In our implementation, we also feed the output of DCNN to WAS and TAS. The total objective to train the new architecture is defined as:
\begin{equation}
    \min_{\theta} \mathcal{L} = \mathcal{L}_{Segment} + \alpha \times \mathcal{L}_{WAS} + \beta \times \mathcal{L}_{TAS},
\end{equation}

\noindent where $\mathcal{L}_{Segment}$ is the original loss used to train DeepLabV3+~\cite{chen2018encoder}, $\mathcal{L}_{WAS}$ and $\mathcal{L}_{TAS}$ are the cross-entropy losses utilized to train WAS and TAS, respectively. $\alpha$ and $\beta$ are scalars to ensure numerical stability during the training and to give more emphasis to the main loss, i.e., $\mathcal{L}_{Segment}$. It should be noted that each loss is back-propagated separately. $\mathcal{L}_{Segment}$ is back-propagated over all the architecture except WAS and TAS. On the other hand, $\mathcal{L}_{WAS}$ and $\mathcal{L}_{TAS}$ are back-propagated only to DCNN.

\noindent\textbf{Synthetic-aware training procedure.}
Training on source domain and fine-tuning on the target domain is a well-known approach to mitigate the domain gap~\cite{tzeng2017adversarial}. However, it is not practical as it requires annotated real data from the target domain which may not be always affordable. Furthermore, training the model on data from one distribution and then forcing the model to learn a new distribution limits the ability of the network to learn and may not converge to a global minima. 

Thus, we propose training our modified DeepLabV3+~\cite{chen2018encoder} on data from both synthetic and real distributions simultaneously and from scratch. For that aim, we train in alternating fashion: one batch from~$\mathcal{D}_{stand-real}$ and next batch from~$\mathcal{D}_{adv-synth}$. At the same time, since the aim is to learn how to extract useful features under adverse conditions, we freeze DCNN weights when training on a batch from~$\mathcal{D}_{stand-real}$ and update them for a batch from~$\mathcal{D}_{adv-synth}$. It is worth noting that all other weights are updated for data from both domains. This strategy encourages the encoder to leverage synthetic data to better learn feature extraction for the target domain while it ensures that the decoder is learning how to interpret both features to perform segmentation task under standard and adverse conditions. 
% --------------------------------------------------------------------------
% 
% Table Res 2 (Synthetic data based approach, fine-tune (last layers only))
\begin{table}
\centering
\caption{{\bf mIoU results for our approach Vs. standard domain adaptation methods.} Training our weather and nighttime-aware architecture on both  Cityscapes~\cite{cordts2016cityscapes} and AWSS, improves the performance on ACDC~\cite{sakaridis2021acdc} dataset and achieves adequate peformance on Cityscapes~\cite{cordts2016cityscapes}. Best results are \textbf{bolded}. Fnt stands for Fine-Tuned.}
\label{table:perf_improv_data_based}
\resizebox{\textwidth}{!}{%
\begin{tabular}{clccccc|c}
\multicolumn{1}{l}{} &  & \multicolumn{5}{c}{ACDC} & Cityscapes \\
\cmidrule(lr{1em}){3-7} \cmidrule(lr{1em}){8-8}
\multicolumn{1}{l}{} &  & Rain & Fog & Snow & Night & Overall & Overall \\
\midrule
\multirow{2}{*}{DeepLabV3+~\cite{chen2018encoder}} & Baseline & 0.41 & 0.46 & 0.36 & 0.17 & 0.35 & 0.78 \\

 & FnT on AWSS & 0.44 & 0.48 & 0.47 & 0.19 & 0.39 & 0.59 \\
\cmidrule(lr{1em}){2-8}
\multirow{2}{*}{HRNet~\cite{yuan2019segmentation}} & Baseline & 0.46 & 0.42 & 0.41 & 0.09 & 0.35 & 0.75 \\

 & FnT on AWSS & 0.47 & 0.49 & 0.35 & 0.14 & 0.36 & 0.51 \\
\cmidrule(lr{1em}){2-8}
\multirow{2}{*}{DANet~\cite{fu2019dual}} & Baseline & 0.47 & 0.57 & 0.44 & 0.21 & 0.42 & 0.82 \\

 & FnT on AWSS & 0.48 & 0.58 & 0.48 & 0.26 & 0.45 & 0.74 \\
\cmidrule(lr{1em}){2-8}
\multirow{2}{*}{PSPNet~\cite{zhao2017pyramid}} & Baseline & 0.49 & 0.54 & 0.43 & 0.20 & 0.41 & \textbf{0.86} \\
 & FnT on AWSS & 0.52 & 0.56 & 0.46 & 0.18 & 0.43 & 0.86\\

\cmidrule[1pt]{2-8}
Ours & Full-Model & \textbf{0.57} & \textbf{0.60} & \textbf{0.50} & \textbf{0.27} & \textbf{0.49} & 0.75
\end{tabular}%
}
% \vspace{-30px}
\end{table}

\section{Experiments}

\paragraph*{Datasets.} For training experiments, we use two datasets: AWSS dataset and the training split of Cityscapes~\cite{cordts2016cityscapes} dataset. For evaluation, we use validation splits of Cityscapes and ACDC~\cite{sakaridis2021acdc} datasets. The three datasets follow the same conventions and color codes. 
Cityscapes comprises 2975 training images and 500 validation images. It is captured in urban scenes under normal weather conditions in the daytime. ACDC validation split comprises 506 images spanning rainy, foggy, snowy weather conditions and nighttime attributes. \\
\textbf{Implementation details.} Experiments are conducted on a Tesla V100 GPU. For all experiments, we keep the default parameters of the authors. For our adopted DeepLabV3+ architectures, we use a batch size of 4 while we keep all other parameters same as DeepLabV3+. For DeepLabV3+ baseline, our architecture, and all ablation study experiments, we train for 30K iterations. We set $\alpha = \beta = 10^{-5}$, as these values achieved the best results. \\
\textbf{Baselines.}
To analyse the robustness of recent semantic segmentation methods under adverse conditions, we use  DeepLabV3+~\cite{chen2018encoder}, HRNet~\cite{yuan2019segmentation}, DANet~\cite{fu2019dual}, and PSPNet~\cite{zhao2017pyramid}. \\
\textbf{Evaluation metric.} We use the common Mean Intersection over Union (mIoU)~\cite{chen2018encoder,yuan2019segmentation,fu2019dual,zhao2017pyramid} on the validation sets of Cityscapes and ACDC similar to~\cite{chen2020dsnet,xie2022towards,musat2021multi}.
\subsection{Results}
Before discussing our architecture results, we will discuss how the domain shift degrades the state-of-the-art, and the improvements achieved by fine-tuning on synthetic data.\\
\textbf{Standard-Adverse domain shift.} As shown by our results in Table~\ref{table:perf_improv_data_based}, the performance of recent methods clearly degrade under adverse weather conditions and at nighttime (see rows \textit{Baseline}). Additionally, it seems that snow and nighttime represent a clear challenge for recent models. Snow causes a drastic change in scene appearance: falling snow particles, snow on pavements and other scene elements makes these objects considerably different compared to what the model learned in the training phase. Thus, the model struggles to segment these elements. Similarly, nighttime scenes with the radical decrease in illumination presents a major challenge for segmentation methods.\\
\textbf{Domain adaptation using synthetic data.}
% Transfer learning is a cheap and direct solution usually applied to handle the domain shift problem. It requires less training data from the target domain. Although it can improve the performance on the target domain, it generally degrades the performance on the source domain.
% Unlike other works that use real data to fine-tune the trained models~\cite{chu2016best,tajbakhsh2016convolutional,guo2019spottune}, we utilize only synthetic data  to alleviate the domain gap. [This should be on related work]
Transfer learning is usually applied to handle a domain shift. However, although it improves the performance on the target domain, it generally degrades the performance on the source domain. As shown in Table ~\ref{table:perf_improv_data_based} (FnT on AWSS), we can improve the performance of each semantic segmentation model.
 For some attributes like night and snow, the improvement was more than 50\% (e.g. HRNet~\cite{yuan2019segmentation} under night). Generally, each semantic segmentation model was able to leverage AWSS to improve its performance for each adverse attribute. However, when evaluating these fine-tuned models on the original domain (Cityscapes), we see a clear degradation in performance. This degradation was more severe for some models like HRNet~\cite{yuan2019segmentation} while it was slight for PSPNet~\cite{zhao2017pyramid}
 
%  This degradation was more sever for some models like HRNet~\cite{yuan2019segmentation} while it was slight for~PSPNet~\cite{zhao2017pyramid}. 

%This can be linked to the architecture of the semantic segmentation model, the optimization technique, and the other hyper-parameters.  % Cut this phrase? Is this implying that perhaps we should have optimised the hyper-parameters to avoid this degradation?

% Table Res 1 Kerim commented this for space and because it is redundant (it was table 2). It is included in now Table 2 (was table 3).
% \begin{table}
% \centering
% \caption{{\bf Segmentation results on Cityscapes~\cite{cordts2016cityscapes} and ACDC~\cite{sakaridis2021acdc}.} State-of-the-art semantic segmentation methods perform well under standard conditions but struggle at adverse conditions.}
% \label{table:perf_deg_acdc}
% % \resizebox{\linewidth*0.8}{!}{%
% \begin{tabular}{cccccc|c}
% \multicolumn{1}{l}{} & \multicolumn{5}{c}{ACDC} & Cityscapes \\
% \cmidrule(lr{1em}){2-6} \cmidrule(lr{1em}){7-7}
% \multicolumn{1}{l}{} & Rain & Fog & Snow & Night & Overall & Overall \\
% \midrule
% DeepLabV3+~\cite{chen2018encoder} & 0.41 & 0.46 & 0.36 & 0.17 & 0.35 & 0.78 \\
% HRNet~\cite{yuan2019segmentation} & 0.46 & 0.42 & 0.41 & 0.09 & 0.35 & 0.75 \\
% DANet~\cite{fu2019dual} & 0.47 & 0.57 & 0.44 & 0.21 & 0.42 & 0.82 \\
% PSPNet~\cite{zhao2017pyramid} & 0.49 & 0.54 & 0.43 & 0.20 & 0.41 & 0.86
% \end{tabular}%
% % }
% \end{table}

% Table Res 3.
\begin{table}[t]
\centering
\caption{{\bf Per-class mIoU results on ACDC~\cite{sakaridis2021acdc} dataset.} Our model achieves the best overall results on ACDC~\cite{sakaridis2021acdc}. It maintains the best results on \textit{Road}, \textit{Sidewalk}, \textit{Building}, and \textit{Person} classes. Best and second best results are \textbf{bolded} and \underline{underlined}, respectively.}
\label{table:per-class_results}
\resizebox{\textwidth}{!}{%
\begin{tabular}{lcccccccccc|c}
           & Road & Sidewalk & Building & Pole & Tr. Light & Tr. Sign & Vegetation & Sky  & Person & Car  & Overall \\
\midrule
DeepLabV3+ & \underline{0.71} & \underline{0.22}     & 0.31     & 0.18 & 0.22      & 0.29     & \textbf{0.72}       & 0.38 & 0.24   & 0.23 & 0.35    \\
HRNet      & 0.55 & 0.16     & 0.44     & 0.14 & 0.28      & 0.24     & 0.66       & \textbf{0.72} & 0.07   & 0.19 & 0.35    \\
DANet      & 0.68 & 0.11     & 0.19     & \underline{0.28} & \textbf{0.54}      & \textbf{0.67}     & 0.26       & 0.65 & \underline{0.29}   & \textbf{0.53} & \underline{0.42}   \\
PSPNet     & 0.63 & 0.12     & \underline{0.60}     & \textbf{0.30} & \underline{0.48}      & \underline{0.41}     & 0.62       & 0.61 & 0.21   & 0.17 & 0.41    \\
Ours  &     \textbf{0.79} &	\textbf{0.40} &	\textbf{0.63} &	0.25 &	0.26 &	0.33 &	\underline{0.69} &	\underline{0.66} &	\textbf{0.32} &	\underline{0.52} &	\textbf{0.49}   
\end{tabular}%
}
\end{table}

\begin{figure*}[t]
    \centering
    \includegraphics[width=\textwidth]{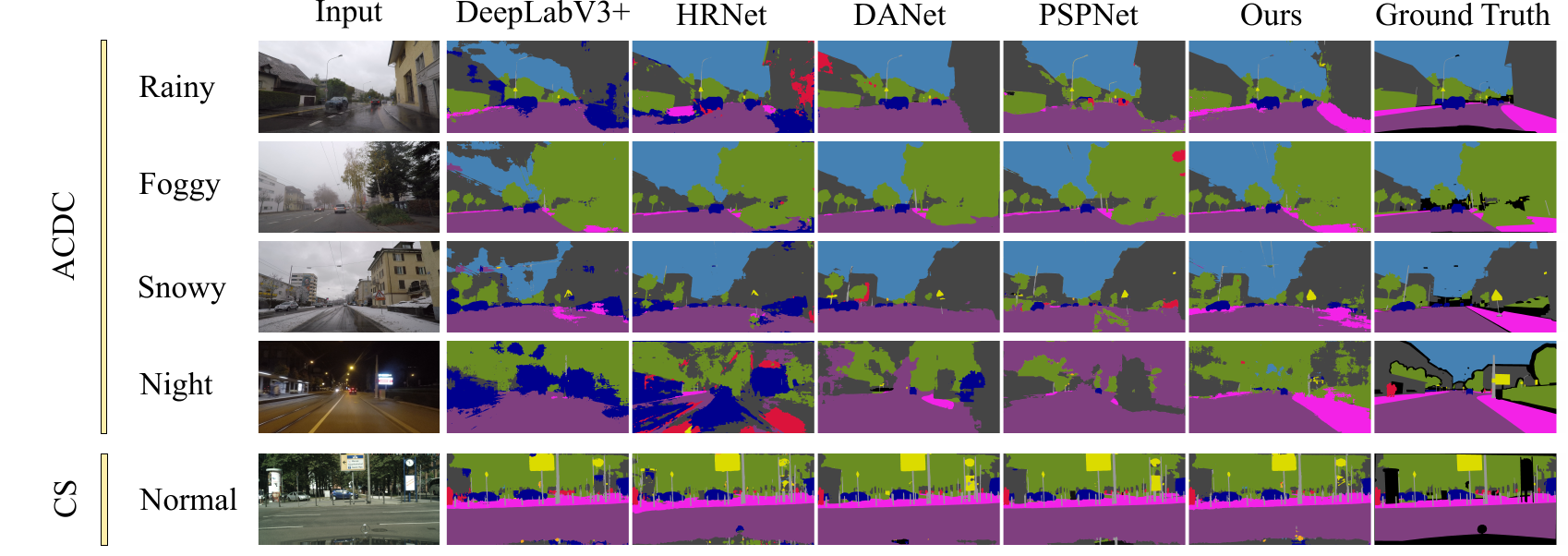}
    \caption{\textbf{Visual comparison between baselines and our approach.} Segmentation results are shown on ACDC~\cite{sakaridis2021acdc} and Cityscapes~\cite{cordts2016cityscapes} dataset, respectively.}
    \label{fig:qualitative}
\end{figure*}

\noindent\textbf{Weather and night aware architecture.}
While the previous solution is simple, the improvement on the target domain was limited, and the performance on the source domain was sharply degraded. As a remedy, our architecture based solution achieves the best results on the target domain and it maintains an adequate performance on the source domain. As reported in Table~\ref{table:perf_improv_data_based}, making the model aware of the weather condition and daytime-nighttime attributes of the images in the training phase helps the model to learn how to extract more efficient features under both standard and challenging scenarios. Qualitative results are shown in Figure~\ref{fig:qualitative}. Furthermore, per-class results are demonstrated in Table~\ref{table:per-class_results}, our model achieves the best results on \textit{Road}, \textit{Sidewalk}, \textit{Building}, and \textit{Person} semantic classes. The largest improvement was on the \textit{Sidewalk} which is around 82\% improvement compared to DeepLabV3+, the best performing baseline on this class. As expected, snow and rain changes the visual appearance of this class significantly.  This is because of snow accumulation, footsteps on snow, rain splash and mud, in addition to light reflection due to wet surface when raining.\\

\subsection{Ablation Study}
To understand the effect of each design decision, we perform several experiments. \\% as detailed below.\\ remove widow
% \textbf{Training data type.} We train the baseline model on AWSS from scratch and report the results in Table~\ref{table:ablation} first row. As expected, training on synthetic data alone does not achieve satisfactory results due to domain gap problem between synthetic and real data. Thus, this experiment suggests that AWSS can be used a complementary to the real data and not as an alternative. On the other hand, training the model from scratch on standard weather condition will perform well on standard conditions but will fail under challenging conditions as shown in Table~\ref{table:ablation} second row.\\
\noindent\textbf{Training data type.} We train the baseline model on AWSS from scratch (Table~\ref{table:ablation} first row). As expected, training on synthetic data alone does not achieve satisfactory results due to domain gap between synthetic and real data. Thus, this suggests that AWSS can be used as complementary to the real data and not as an alternative. On the other hand, training the model from scratch on standard weather will perform well on these conditions but will fail under challenging conditions (Table~\ref{table:ablation} second row).\\
% \textbf{Training strategy.} As shown in Table~\ref{table:ablation} third row, the standard method of transfer learning i.e., fine-tuning the last layers of the model on the target domain can improve the performance on the target domain but degrades the performance on the source domain. Thus, it may not seem as a legitimate solution.\\
\textbf{Training strategy.} As shown in Table~\ref{table:ablation} third row, the standard method of transfer learning (fine-tuning the last layers on the target domain) improves the performance on the target domain but degrades the performance on the source domain.\\
\textbf{Weather-Time awareness.} 
Our approach achieves the best results under adverse conditions while still maintaining a satisfactory performance under standard conditions. Making the model synthetic aware and training the model without weather and nighttime-awareness achieve better results on the source domain but low performance on the target domain, compared to fine-tuning experiment. Adding the weather awareness to the model, i.e WAS,  improves the performance at standard and adverse conditions. All adverse weather attributes were improved clearly as expected but the night attribute maintained a slight improvement. Finally, making the model aware of nighttime too, boosts significantly the performance under nighttime. Interestingly, it improves also the performance of the other weather conditions too. This is expected as TAS and WAS teachers allow the model to learn weather specific and nighttime-specific robust features which enables the model to achieve better results under these challenging conditions. 

% Table~\ref{table:ablation} demonstrates the effect of each design option. 
% Table Ablation Weather and Daytime Awareness  
\begin{table}

\centering
% \caption{{\bf Ablation analysis of weather and time awareness on performance.} Making the DeepLabV3+ weather and time aware improved the performance significantly at both normal weather i.e. Cityscapes~\cite{cordts2016cityscapes} (CS)  and adverse weather i.e. ACDC~\cite{sakaridis2021acdc} scenarios. Best and second best results are \textbf{bolded} and \underline{underlined}, respectively.}
\caption{{\bf Ablation analysis of weather and time awareness on performance.} Making the DeepLabV3+ weather and time aware improved the performance significantly at both normal weather, i.e. Cityscapes~\cite{cordts2016cityscapes} (CS), and adverse weather, i.e. ACDC~\cite{sakaridis2021acdc}, scenarios. Best and second best results are \textbf{bolded} and \underline{underlined}, respectively.}
\label{table:ablation}
\resizebox{\textwidth}{!}{%
% \begin{tabular}{cccccc|c}
% \multicolumn{1}{l}{}                                     & \multicolumn{5}{c}{ACDC}         & \multicolumn{1}{|c}{Cityscapes} \\
% \cmidrule(lr{1em}){2-6} \cmidrule(lr{1em}){7-7}
% \multicolumn{1}{l}{} &
%   \multicolumn{1}{c}{Rain} &
%   \multicolumn{1}{c}{Fog} &
%   \multicolumn{1}{c}{Snow} &
%   \multicolumn{1}{c}{Night} &
%   \multicolumn{1}{c|}{Overall} &
%   \multicolumn{1}{c}{Overall} \\
%   \midrule
%  Baseline trained from scratch on AWSS                      & 0.24 &	0.25 &	0.26 &	0.11 &	0.22 &	0.27 \\
% Baseline trained from scratch on CS                      & 0.41 & 0.46 & 0.36 & 0.17 & 0.35 & \textbf{0.78} \\
% \midrule
% Trained on CS and fine-tuned on AWSS                     & 0.44 & 0.48 & 0.47 & 0.19 & 0.39 & 0.59          \\
% \midrule
% Ours trained from scratch on CS and AWSS                 & 0.41 & 0.43 & 0.38 & 0.19 & 0.35 & 0.69          \\
% % \midrule
% Ours trained from scratch on CS and AWSS + Weather Aware & \underline{0.49} & \underline{0.55} & \underline{0.47} & \underline{0.20} & \underline{0.43} & \textit{0.73}          \\
% % \midrule
% Ours trained from scratch on CS and AWSS + Weather and Daytime Aware & \textbf{0.57} & \textbf{0.60} & \textbf{0.50} & \textbf{0.27} & \textbf{0.49} & \underline{0.75}
% \end{tabular}%

\begin{tabular}{l|cccccc|c}
 &
  \multirow{2}{*}{Training Mode} &
  \multicolumn{5}{c|}{ACDC} &
  Cityscapes \\
 &
   &
  Rain &
  Fog &
  Snow &
  Night &
  Overall &
  Overall \\ \cline{1-8} 
\multicolumn{1}{c|}{\multirow{3}{*}{Baseline}} &
  scratch on AWSS &
  0.24 &
  0.25 &
  0.26 &
  0.11 &
  0.22 &
  0.27 \\
\multicolumn{1}{c|}{} &
  scratch on CS &
  0.41 &
  0.46 &
  0.36 &
  0.17 &
  0.35 &
  \textbf{0.78} \\ \cline{2-8} 
\multicolumn{1}{c|}{} &
  scratch on CS and fine-tuned on AWSS &
  0.44 &
  0.48 &
  0.47 &
  0.19 &
  0.39 &
  0.59 \\ \cline{1-8} 
\multicolumn{1}{l|}{\multirow{3}{*}{Ours}} &
  scratch on CS and AWSS &
  0.41 &
  0.43 &
  0.38 &
  0.19 &
  0.35 &
  0.69 \\
\multicolumn{1}{l|}{} &
  scratch on CS and AWSS + Weather Aware &
  \underline{0.49} &
  \underline{0.55} &
  \underline{0.47} &
  \underline{0.20} &
  \underline{0.43} &
  \textit{0.73} \\
\multicolumn{1}{l|}{} &
  scratch on CS and AWSS + Weather and Nighttime Aware &
  \textbf{0.57} &
  \textbf{0.60} &
  \textbf{0.50} &
  \textbf{0.27} &
  \textbf{0.49} &
  \underline{0.75}
\end{tabular}%
}
\end{table}

%------------------------------------------------------------------------
\section{Conclusions}
We introduce a novel synthetic dataset, the AWSS, that covers various adverse conditions. We show that fine-tuning four state-of-the-art semantic segmentation models improve performance under adverse conditions but degrades the performance under standard conditions. Our proposed solution shows that making the model aware of the synthetic data and utilizing weather-aware-supervisor and time-aware-supervisor achieves the best results under adverse weather conditions while maintaining an adequate performance under standard conditions. 
\section*{Acknowledgement}
This work was funded by the Faculty of Science and Technology of Lancaster University. We thank the High End Computing facility of Lancaster University for the computing resources. The authors would also like to thank CAPES, CNPq, and FAPEMIG for funding different parts of this work.
% CAPES (#88881.120236/2016-01)
\FloatBarrier
\bibliography{egbib}
\end{document}